\documentclass[10pt,journal,compsoc]{IEEEtran}

\usepackage{cite}
\usepackage{url}
\usepackage{ragged2e}
\usepackage{epsfig}
\usepackage{graphicx}
\usepackage{amsmath,amssymb} 
\usepackage{subfigure}
\usepackage{algorithm}
\usepackage{algorithmicx}
\usepackage{algpseudocode}
\usepackage{graphics}
\usepackage{threeparttable}
\usepackage{color}
\usepackage[normalem]{ulem}
\usepackage{multirow}
\usepackage{float}
\usepackage{amsfonts}
\usepackage{bm}
\usepackage{array}
\usepackage[table]{xcolor}
\usepackage{colortbl}
\usepackage{pifont}
\usepackage{diagbox}
\usepackage{rotating}
\usepackage{booktabs}
\usepackage{overpic}
\usepackage{textcomp}
\usepackage{contour}
\usepackage{enumitem}
\usepackage[colorlinks=true]{hyperref}

\RequirePackage{silence}
\definecolor{bblue}{rgb}{0,150,230}
\definecolor{mygray}{gray}{.92}

\newcommand{\secref}[1]{Section \ref{#1}}

\graphicspath{{./Imgs/}}
\DeclareGraphicsExtensions{.jpg,.pdf,.png}

\makeatletter
\def\ps@IEEEtitlepagestyle{%
  \def\@oddfoot{\mycopyrightnotice}%
  \def\@evenfoot{}%
}
\def\mycopyrightnotice{%
  {\hfill \scriptsize {This work has been submitted to the IEEE for possible publication.
  Copyright may be transferred without notice, after which this version may no longer be accessible.}\hfill}

}
\makeatother

\begin{document}

\title{Knowledge-enriched Attention Network with Group-wise Semantic for Visual Storytelling}

\author{Tengpeng~Li,~\IEEEmembership{}
        Hanli~Wang,~\IEEEmembership{Senior Member,~IEEE,}
        Bin~He,~\IEEEmembership{}
        Chang~Wen~Chen,~\IEEEmembership{Fellow,~IEEE}

        \thanks{
         {Corresponding author: Hanli~Wang.}
         }

         \thanks{{T.~Li and H.~Wang are with the Department of Computer Science \& Technology, Key Laboratory of Embedded System and Service Computing (Ministry of Education), Tongji University, Shanghai 200092, P. R. China, and with Shanghai Research Institute for Intelligent Autonomous System, Shanghai 201210, P. R. China (e-mail: ltpfor1225@tongji.edu.cn, hanliwang@tongji.edu.cn).}}

         \thanks{{B.~He is with Shanghai Research Institute for Intelligent Autonomous System, Shanghai 201210, P. R. China (e-mail: hebin@tongji.edu.cn).}}


         \thanks{{C. W. Chen is with the Department of Computing, Hong Kong Polytechnic University, Hung Hom, Kowloon, Hong Kong SAR, China (e-mail: Changwen.chen@polyu.edu.hk).}}
}


\IEEEtitleabstractindextext{%
\begin{abstract}
\justifying
As a technically challenging topic, visual storytelling aims at generating an imaginary and coherent story with narrative multi-sentences from a group of relevant images. Existing methods often generate direct and rigid descriptions of apparent image-based contents, because they are not capable of exploring implicit information beyond images. Hence, these schemes could not capture consistent dependencies from holistic representation, impairing the generation of reasonable and fluent story. To address these problems, a novel knowledge-enriched attention network with group-wise semantic model is proposed. Three main novel components are designed and supported by substantial experiments to reveal practical advantages. First, a knowledge-enriched attention network is designed to extract implicit concepts from external knowledge system, and these concepts are followed by a cascade cross-modal attention mechanism to characterize imaginative and concrete representations. Second, a group-wise semantic module with second-order pooling is developed to explore the globally consistent guidance. Third, a unified one-stage story generation model with encoder-decoder structure is proposed to simultaneously train and infer the knowledge-enriched attention network, group-wise semantic module and multi-modal story generation decoder in an end-to-end fashion. Substantial experiments on the popular Visual Storytelling dataset with both objective and subjective evaluation metrics demonstrate the superior performance of the proposed scheme as compared with other state-of-the-art methods.
\end{abstract}
}

\maketitle
\IEEEdisplaynontitleabstractindextext

\IEEEpeerreviewmaketitle

\IEEEraisesectionheading{\section{Introduction}
\label{sec:introduction}}

Visual storytelling, which aims at producing a set of expressive and coherent sentences to depict the contents of a group of sequential images, has been an interesting and rapidly growing research topic in the fields of computer vision and multimedia computing. Different from visual captioning which devotes to describe the superficial contents in an image or a video, visual storytelling is expected not only to recognize the diverse semantical contexts and relations within one image and across images, but also to generate the storyline of image stream and express more implicit imaginations out of the images. Visual storytelling can be used in many real-world applications, such as helping the disabled to comprehend image contexts from social media, verifying advanced properties of intelligent devices, etc.

In visual storytelling, it is essential to learn the storyline and express with informative sentences. Therefore, valuable contextual information should be deduced for the target image stream. In general, a visual storytelling model intends to solve two main issues:
(1) generating the abundant information of extracted features in single image, and (2) providing the precise storyline about the event occurred in the image sequence. On one hand, most visual captioning schemes focus on detecting visual features, where convolutional features~\cite{wang2018no, huang2019attention, jung2020hide} and object features~\cite{pan2020x, cornia2020meshed, luo2021dual} have been widely used in these schemes. Nevertheless, regional-visual features can merely detect the intrinsic and superficial information, lacking the the capability to explore diverse and creative textures that were not apparent from images. Several recent approaches~\cite{hsu2020knowledge, chen2021commonsense, yang2019auto, zhong2020comprehensive} introduced external knowledge by leveraging graph-based structures like the scene graph~\cite{xu2017scene} and the commonsense graph~\cite{speer2017conceptnet} to strengthen symbolic creativity and achieve desired performances. Nonetheless, these approaches either did not establish the associations of cross-modal information or only learned the implicit external contents in two separated stages, leading to sub-optimal performance. We strongly believe that the attentive visual and textual representations are essential to produce concrete and imaginative descriptions.

On the other hand, a number of unified frameworks~\cite{wang2018no, jung2020hide, wang2020storytelling} have been developed recently to solve the problem of lacking global consistency in image sequence, where the recurrent neural network~(RNN)~\cite{wang2018no, jung2020hide} or temporal convolutional network~(TCN)~\cite{wang2020storytelling} has been adopted to explore the temporal feature relations. However, both RNN and TCN encounter problems in their optimization~\cite{vaswani2017attention} because of memory dilution along the longer feature sequence, failing to generate the topic-aware information of an image stream. Nevertheless, the storyline containing long-range dependencies is crucial to output the coherent multi-sentences. Furthermore, the most serious problem among existing approaches is that they are incapable of establishing a unified framework to simultaneously capture sufficient regional features and topic-aware global features for visual storytelling.

To address the aforementioned challenges, a knowledge-enriched attention network with group-wise semantic~(KAGS) model is proposed in this research for visual storytelling. The proposed KAGS model will first leverage a CNN~\cite{he2016deep} and a Faster-RCNN~\cite{ren2016faster} as encoder to extract convolutional features, semantic labels and regional object features from the input image stream. Then the semantic labels and regional features will be sent into the proposed knowledge-enriched attention network~(KAN), where the semantic labels are processed with ConceptNet~\cite{speer2017conceptnet} and the regional features are dealt with the cascade cross-modal attention module. The proposed KAN can achieve sufficient feature representation to enable the establishment of cross-modal correlations of both textual and visual information. Meanwhile, the group-wise semantic module~(GSM) with second order pooling~(SOP) is introduced to transform the convolutional group features into global guided vector. Different from the sequential memory enhanced behaviour in RNN or TCN, GSM directly computes the higher-order interaction of any local or non-local pairwise convolutional vectors, in spite of their intra- or inter-spatial positions. The designed GSM can obtain the global feature guidance because it can capture the long-range dependencies of the sequential convolutional features. Finally, the optimized visual and textual features, combined with the global semantic vector, are sent into a multi-modal story decoder to generate the story. As a result, a unified one-stage framework with superior performance is established to optimize all proposed modules for attentive cross-modal features and global semantic guidance in an end-to-end manner.
Major contributions of this work are summarized below.
\begin{itemize}

\item A knowledge-enriched attention network is designed to capture attentive enriched contexts and visual representations to address the problem in external information shortage and feature distraction. The contexts are generated from commonsense graphs and the cascade cross-modal attention is employed to highlight the valuable embedding of heterogeneous information. \vspace{0.2cm}

\item A group-wise semantic module is developed to capture the global consistency of an image stream to overcome the challenge about the incoherent descriptions in a story. This module is able to compute the higher-order interaction of any pairwise semantic vectors regardless of spatial distance restriction, thus contributing to the accurate guidance of the storyline. \vspace{0.2cm}

\item A unified one-stage visual storytelling framework with encoder-decoder structure
is devised to simultaneously optimize the knowledge-enriched attention network, group-wise semantic module and multi-modal story decoder in an end-to-end fashion. It has been shown that the proposed KAGS scheme is both efficient and effective.

\end{itemize}

The rest of this paper is organized as follows. We introduce in ~\secref{sect:rlwk} the related works in both image captioning and visual storytelling. The proposed knowledge-enriched attention network with group-wise semantic model is described in detail in ~\secref{sect:prop}. We present in ~\secref{sect:exp} the statistic performances, ablative studies and visualization analyses. Finally, we conclude this paper in ~\secref{sect:cln}.

\section{Related Work}
\label{sect:rlwk}

\subsection{Image Captioning}
\label{sect:rlwk:ic}

Image captioning aims at automatically generating a natural language sequence to depict the complex visual contents occurred in a single image, and it can be generally divided into two categories. First, benefiting from the rapid developing technology of natural language machine translation, most early approaches~\cite{mao2014explain, vinyals2015show, jia2015guiding, karpathy2015deep, donahue2015long, rennie2017self, liu2018simnet} attempted to establish the captioning framework with encoder-decoder structure and achieved satisfying performances. In these common approaches, CNN was usually regraded as encoder to extract image features, and RNN was often used to decode the integrated representations for sentence production. In~\cite{mao2014explain}, Mao~\textit{et al.} designed the m-RNN framework consisted of two sub-networks including a CNN-based image encoder and a RNN-based sentence decoder to accomplish sentence generation. Vinyals~\textit{et al.}~\cite{vinyals2015show} leveraged the CNN to extract visual representations and applied the long short-term memory~(LSTM)~\cite{hochreiter1997long} to output the final image description. Jia~\textit{et al.}~\cite{jia2015guiding} proposed a gLSTM model to add the extracted semantic contexts in each LSTM unit for guiding global image content generation. Second, a set of innovated methods with an attention mechanism~\cite{xu2015show,you2016image,lu2017knowing,li2019know,yang2020captionnet} have been proposed to further improve image captioning performances by highlighting meaningful visual and textual information in recent years. Xu~\textit{et al.}~\cite{xu2015show} designed a LSTM-based decoder with soft attention and hard attention modules to focus on important image areas for generating accurate words in the decoding process. You~\textit{et al.}~\cite{you2016image} presented a semantic attention model that integrates the extracted semantic visual feature proposals into the hidden states and RNN-based decoders for better language description. In~\cite{lu2017knowing}, Lu~\textit{et al.} developed an adaptive attention structure to selectively choose image regions for obtaining meaningful features. Furthermore, Anderson~\textit{et al.}~\cite{anderson2018bottom} proposed a bottom-up and top-down attention framework to explore the object-level salient regions and relate each region with one corresponding word for sentence generation. Li~\textit{et al.}~\cite{li2019know} performed a scene graph strategy~\cite{xu2017scene} to capture enriched structural information with semantic entities and pairwise relations. Yang~\textit{et al.}~\cite{yang2020captionnet} proposed the CaptionNet model as an enhanced LSTM to focus on positive visual cues and absorb richer semantics for better feature encoding. In this work, the encoder-decoder structure is also developed by additionally merging attention mechanism and global guidance for robust feature representation.

\begin{figure*}[htbp]
\begin{center}
\includegraphics[width=1.0\linewidth]{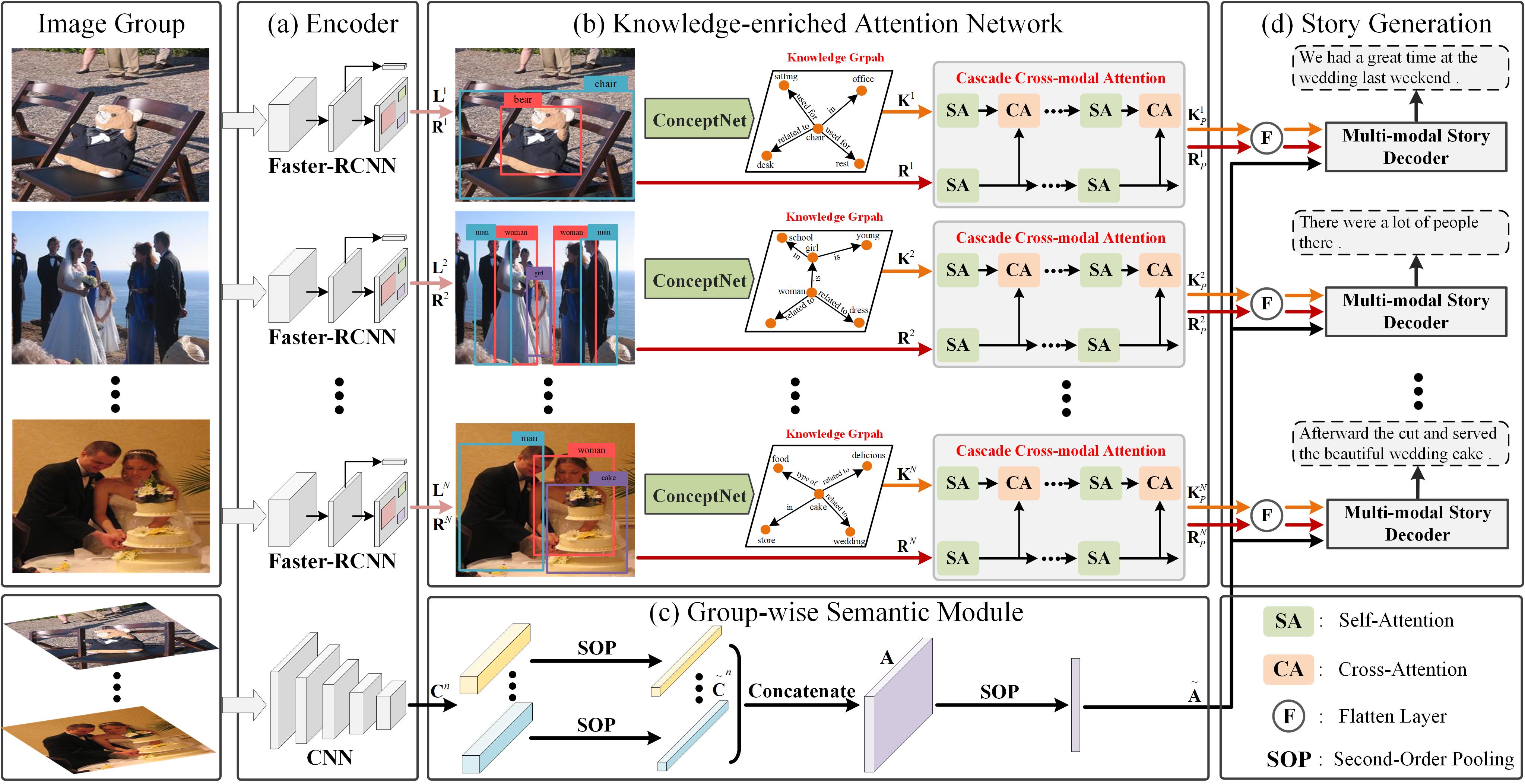}
\end{center}
   \caption{Pipeline of the proposed KAGS for visual storytelling. The framework contains four key components: (a) a Faster-RCNN network and a ResNet backbone to extract regional features and high-level convolutional features; (b) the proposed KAN to obtain the attentive heterogeneous representations by exploiting the intra- and inter- interactions of visual and knowledge concepts; (c) the proposed GSM to explore the global guided aggregation with a set of hierarchical second-order pooling algorithms in a convolutional feature group; and (d) a story generation that fuses the multi-modal information in a decoder to produce the final predicted sentences.}
\label{fig:Pipeline-acm}
\end{figure*}

\subsection{Visual Storytelling}
\label{sect:rlwk:vs}

Visual storytelling is a challenging task in multimedia communities since the designed approaches should bridge an association between the group of visual messages and the sequential natural languages. As an emerging and promising topic, visual storytelling has attracted much attention of researchers and a number of elaborate innovations are proposed. Generally, visual storytelling models can be grouped into end-to-end framework and multi-stage based approach. First, end-to-end framework is popular due to its efficiency for generating stories in a unified structure. Wang~\textit{et al.}~\cite{wang2018no} proposed a classical visual storytelling framework that has been the most popular base structure of following studies. This framework designed an end-to-end structure to encode the sequential converted features jointly by bidirectional gated recurrent units~(GRU) and decoded these processed features separately for the final story. Huang~\textit{et al.}~\cite{huang2019hierarchically} designed a hierarchical two-level decoder to produce the semantic topic and generate a sentence for each single image, and reinforcement learning was applied for optimization. In~\cite{yang2019knowledgeable}, a commonsense-driven generator was employed to caption essential external messages for abundant multi-sentence expressions. Jung~\textit{et al.}~\cite{jung2020hide} proposed a hide and tell model to acquire the imaginative storyline by bridging the feature gap of image stream. To ensure the interesting and informative characteristics of story, Hu~\textit{et al.}~\cite{hu2020makes} designed three human-like criteria combined with a reinforcement learning structure and achieved superior performances on human evaluation metrics. Second, many multi-stage approaches were also emerging which strengthened the diversity and informativeness of frameworks. Hsu~\textit{et al.}~\cite{hsu2019visual} merged various extracted concepts into decoder for more diverse descriptions. Yao~\textit{et al.}~\cite{yao2019plan} designed a hierarchical framework to plan the storyline in the first stage and wrote the topic-based story in the second stage. Moreover, several works~\cite{hsu2020knowledge, chen2021commonsense} introduced the external commonsense knowledge from bases like OpenIE~\cite{angeli2015leveraging}, Visual Genome~\cite{krishna2017visual} or ConceptNet~\cite{speer2017conceptnet} for more diverse descriptions, where Hsu~\textit{et al.}~\cite{hsu2020knowledge} proposed a three-stage framework to produce external knowledge to guide the decoder, Chen~\textit{et al.}~\cite{chen2021commonsense} designed a concept selection module to select enriched concept candidates and then sent them in a visual-language pre-trained model to produce full stories. In this work, an end-to-end model is designed while considering efficiency, informativeness and coherency. Particularly, the proposed one-stage model can train and inference all modules in a unified fashion to promote its efficiency, and the attentive commonsense knowledge and global semantic are also introduced to increase the feature representation for improving the informativeness and consistency of KAGS, respectively.

\section{Knowledge-enriched Attention Network with Group-wise Semantic}
\label{sect:prop}

\subsection{Framework Overview}
\label{sect:prop:ov}

The proposed KAGS is illustrated in Fig.~\ref{fig:Pipeline-acm}. First, a knowledge-enriched attention network is designed to explore the intra- and inter- interactions of visual and textual features in~\secref{sect:prop:kan}. Meanwhile, a group-wise semantic module with a set of second-order pooling algorithms is developed to capture the global guided aggregation of sequential convolutional features in~\secref{sect:prop:gsm}. Finally, the produced multi-modal features are sent into the multi-modal story decoder to generate the final reasonable and coherent story in~\secref{sect:prop:decoder}.

With a group of $N$ associated images $\mathcal{I}=\{\textbf{\textit{I}}^n\}_{n=1}^N$ as input, the task of visual storytelling aims to exploit the effective intra- and inter-feature representations of this image stream, producing a reasonable and coherent story with multiple descriptive sentences $\mathcal{S}=\{\textbf{S}^n\}_{n=1}^N$. To tackle this issue, a novel KAGS model is elaborately designed to generate the story $\mathcal{S}$ in an end-to-end manner.

The overall structure of the proposed KAGS model is illustrated in Fig.~\ref{fig:Pipeline-acm}, which consists of four main components: (a) Encoder, (b) Knowledge-enriched Attention Network~(KAN), (c) Group-wise Semantic Module~(GSM), and (d) Story Generation. Specifically, given a group of relevant images $\mathcal{I}$, the model first leverages the general object detection framework Faster-RCNN~\cite{ren2016faster} and ResNet~\cite{he2016deep} backbones as the encoder to extract boxes of regional-visual features $\mathcal{R}=\{{\textbf{R}}^n\}_{n=1}^{N}$ and the corresponding labels $\mathcal{L}=\{{\textbf{L}}^n\}_{n=1}^{N}$ with high confidence, and the high-level representations in the last convolutional layer $\mathcal{C}=\{{\textbf{C}}^n\}_{n=1}^{N}$, respectively. Then, $\textbf{L}^{n}$ of each image is fed into KAN to explore external knowledge. For the semantic label ${\textbf{L}}^n\in \mathcal{L}$, the ConceptNet~\cite{speer2017conceptnet} is introduced to generate the knowledge concepts $\textbf{K}^{n}$ from external enhanced knowledge base that can further boost the capability of absorbing imaginative and reasonable concepts, thus a group of knowledge concepts $\mathcal{K}=\{{\textbf{K}}^n\}_{n=1}^{N}$ can be acquired. Moreover, to fully utilize the regional-visual features $\textbf{R}^{n}$ and the knowledge concepts $\textbf{K}^{n}$, a cascade cross-modal attention~(CCA) module is designed to progressively model the dense semantic interactions of intra features (image-to-image or text-to-text) and inter features (image-to-text), outputting the enhanced knowledge concepts and attentive regional-visual features. The whole process is defined as $[\textbf{K}^{n}_{P}, \textbf{R}^{n}_{P}]= \mathcal{F}_{cca}(\textbf{K}^{n}, \textbf{R}^{n})$, where $\mathcal{F}_{cca}(\cdot,\cdot)$ and $P$ represent the function of CCA module and the number of cascade layers in CCA, respectively.

Moreover, the recent works~\cite{gao2019global, pan2020x} have shown that an outstanding non-linear feature capability of second-order pooling is achieved by exploiting both channel-wise and spatial-wise interactions. Thereby, the GSM with hierarchical second-order pooling is designed to capture the topic-aware consistency of group convolutional features $\mathcal{C}=\{{\textbf{C}}^n\}_{n=1}^{N}$, and then produces a global-visual aggregation $\tilde{\textbf{A}}=\mathcal{F}_{gsm}(\mathcal{C})$, which can help to capture the global guided semantic and avoid noisy interference. Finally, the model feds $\textbf{K}^{n}_{P}$, $\textbf{R}^{n}_{P}$ and  $\tilde{\textbf{A}}$ into the multi-modal story decoder, generating the predicted sentence $\textbf{S}^{n}$.

\subsection{Knowledge-enriched Attention Network}
\label{sect:prop:kan}

As aforementioned, to overcome the problem of insufficient external information and distracted features, the knowledge-enriched attention network~(KAN) is designed to increase the external priors from current knowledge repository and establish intra- and inter- dense correlations of cross-modal features. In fact, several existing knowledge-based methods~\cite{wang2020storytelling,yang2020creative,chen2021commonsense} for visual storytelling also devote to leverage external implicit knowledge for better model performance, but they only focus on the intra correspondence of textual concepts instead of considering the inter association of heterogeneous information that is crucial to visual storytelling, resulting in sub-optimal representation capability. Differently, the proposed KAN constructs the interactions of both enriched knowledge and visual concepts based on CCA, which establishes the long-range dependencies of homogeneous and heterogeneous features between any pairwise feature vectors. Therefore, enriched textual knowledge and visual features can be assigned with higher attention weights in meaningful feature dimensions, facilitating to a more optimized visual storytelling estimation than the methods only considering textual information~\cite{wang2020storytelling,yang2020creative,chen2021commonsense}.

\begin{figure}[htbp]
\begin{center}
\includegraphics[width=0.85\linewidth]{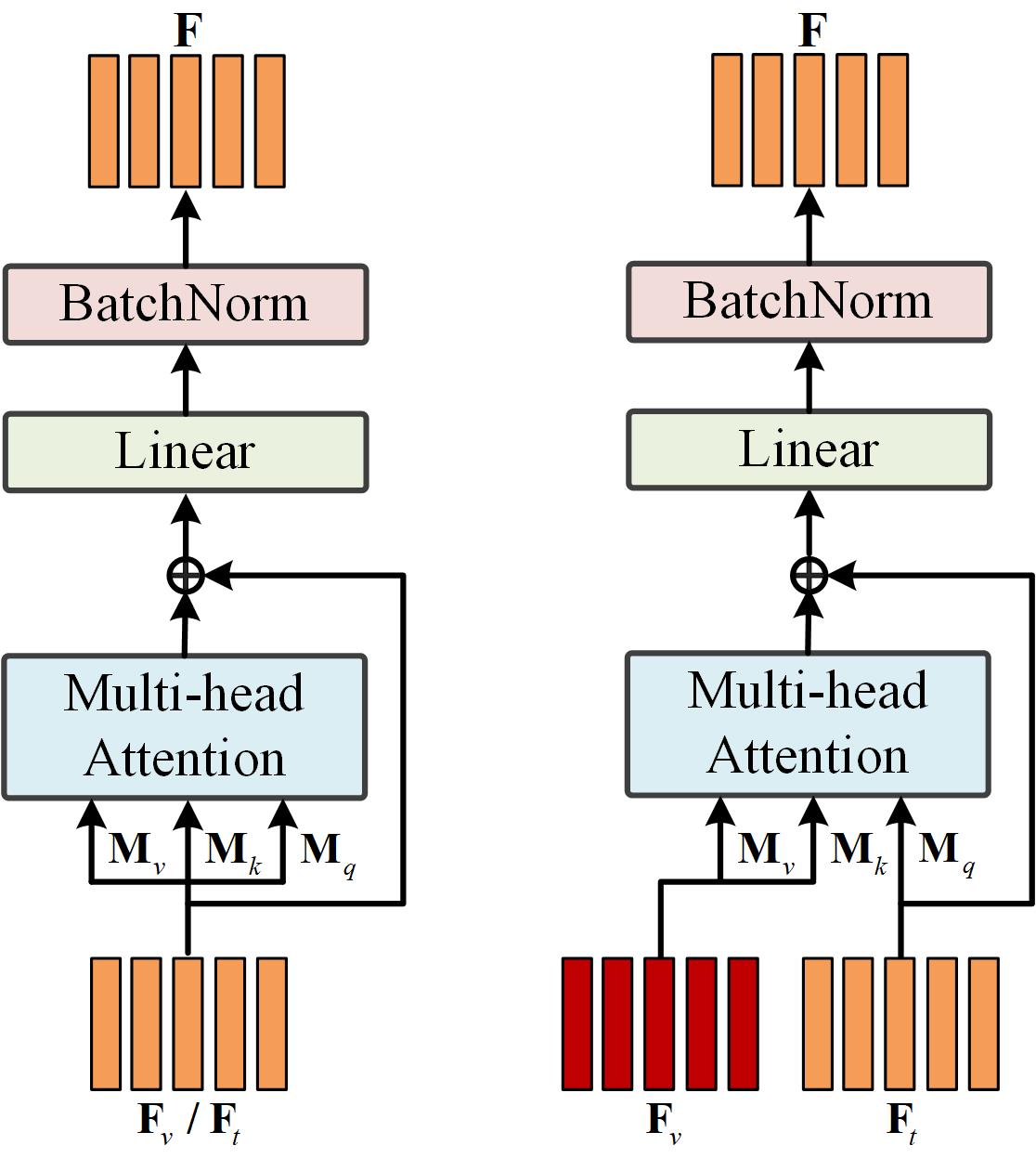}
\end{center}
   \caption{The schematic diagram of two attention units employed by the proposed CCA module, where the left unit is self-attention and the right unit is cross-attention.}
\label{fig:cca}
\end{figure}

\noindent\textbf{Knowledge Graph.} To offer current storytelling datasets more imaginary and reasonable concepts, the proposed KAGS establishes commonsense knowledge graphs based on the semantic labels $\mathcal{L}$ detected by Faster-RCNN~\cite{ren2016faster}. Similar to~\cite{chen2021commonsense,yang2019knowledgeable}, KAGS adopts the generalized ConceptNet~\cite{speer2017conceptnet} as the knowledge extractor to collect numerous commonsense words with rich imagination, abundant emotions and objective facts. Specifically, the knowledge concepts $\textbf{K}^{n}=\{{\textbf{K}}^n_{k}\}_{k=1}^{K}$ is constructed for the given semantic label ${\textbf{L}}^n$, where $K$ indicates to employ the the top-$K$ candidates of the $n^{th}$ image based on their scores of confidence, and each ${\textbf{K}}^n_{k}$ is composed of two entities and one edge relation.

\noindent\textbf{Cascade Cross-modal Attention.} Given the extracted rich knowledge, a tricky challenge is that many selected concepts are irrelevant to the visual information, thus introducing many interferences that reduce the story description accuracy. Recently, the method~\cite{yang2019auto} investigates the visual-textual guided encoding pattern to selectively highlight the positive information and suppress the negative message. Motivated by this and the self-attention mechanism in~\cite{vaswani2017attention}, the CCA module is designed through stacking self-attention~(SA) and cross-attention~(CA) as shown in Fig.~\ref{fig:cca} to progressively explore and optimize cross-modal interactions.
In detail, having the query matrix $\textbf{M}_{q}\in \mathbb{R}^{m\times d}$, the key matrix $\textbf{M}_{k}\in \mathbb{R}^{m\times d}$ and the value matrix $\textbf{M}_{v}\in \mathbb{R}^{m\times d}$, the attentive feature $\textbf{F}\in \mathbb{R}^{m\times d}$ can be obtained by summing all values of $\textbf{M}_{v}$ with the corresponding matrix weights learned from $\textbf{M}_{q}$ and $\textbf{M}_{k}$, and the dot-product attention is defined as
\begin{equation}
\label{eq:1}
\textbf{F}=Attention(\textbf{M}_{q}, \textbf{M}_{k}, \textbf{M}_{v})=softmax(\frac{ \textbf{M}_{q}  \textbf{M}_{k}^\top}{\sqrt{d}}) \textbf{M}_{v},
\end{equation}
where $\frac{1}{\sqrt{d}}$, $m$ and $d$ represent scale factor, vector number and feature dimension, respectively.

In order to enhance the feature capacity of different subspaces, a multi-head attention mechanism~\cite{vaswani2017attention} is also leveraged, which consists of $h$ parallel subspaces. The attentive feature \textbf{F} is formulated as
\begin{equation}
\begin{split}
\label{eq:2}
\textbf{F}&=MultiHead(\textbf{M}_{q}, \textbf{M}_{k}, \textbf{M}_{v})\\
          &= [head^{1}, head^{2}, \cdots, head^{h}] \textbf{W}_{o},
\end{split}
\end{equation}

\begin{equation}
\label{eq:3}
head^{i}=Attention(\textbf{M}_{q} \textbf{W}^{i}_{q}, \textbf{M}_{k} \textbf{W}^{i}_{k}, \textbf{M}_{v}  \textbf{W}^{i}_{v}),
\end{equation}
where $\textbf{W}^{i}_{q} \in \mathbb{R}^{d\times d_{q}}$, $\textbf{W}^{i}_{k} \in \mathbb{R}^{d\times d_{k}}$ and $\textbf{W}^{i}_{v} \in \mathbb{R}^{d\times d_{v}}$ are the learnable projection matrices of the $i^{th}$ head, and $\textbf{W}_{o} \in \mathbb{R}^{(h\times d_{v}) \times d}$. In this schema, the multi-head attention is applied to both of the SA and CA units, followed by the function $LS(\cdot)$ consisting of a point-wise addition, a linear layer and a BatchNorm layer. In Fig.~\ref{fig:cca}, given the visual features $\textbf{F}_{v}$ or the textual features $\textbf{F}_{t}$ of each image, the SA unit outputs the self-attentive representation as
\begin{equation}
\begin{split}
\label{eq:4}
&SA(\textbf{F}_{v})=LS(MultiHead(\textbf{F}_{v}, \textbf{F}_{v}, \textbf{F}_{v})), \\
&SA(\textbf{F}_{t})=LS(MultiHead(\textbf{F}_{t}, \textbf{F}_{t}, \textbf{F}_{t})).
\end{split}
\end{equation}
Similarly, both visual features $\textbf{F}_{v}$ and textual features $\textbf{F}_{t}$ can be fed into CA unit, generating the cross-attentive representation as
\begin{equation}
\label{eq:5}
CA(\textbf{F}_{t}, \textbf{F}_{v})=LS(MultiHead(\textbf{F}_{t}, \textbf{F}_{v}, \textbf{F}_{v})),
\end{equation}

Now, the proposed CCA can be constructed by cascading $P-1$ layers as shown in Fig.~\ref{fig:Pipeline-acm} (b),
which is represented as $\mathcal{F}_{cca}=[\mathcal{F}_{cca}^{(1)}, \mathcal{F}_{cca}^{(2)}, \cdots, \mathcal{F}_{cca}^{(P-1)}]$. Specifically, the $p^{th}$ cascade layer of $\mathcal{F}_{cca}$ including two SA units and one CA unit can be defined as
\begin{equation}
\begin{split}
\label{eq:6}
[\textbf{K}^n_{p+1}, \textbf{R}^n_{p+1}]&=\mathcal{F}_{cca} ^ {(p)}(\textbf{K}^n_{p}, \textbf{R}^n_{p})\\
&=[CA(  SA(\textbf{K}^n_{p}),  SA(\textbf{R}^n_{p}) ),  SA(\textbf{R}^n_{p}) ],
\end{split}
\end{equation}
where $\textbf{K}^{n}_{p}$, $\textbf{R}^{n}_{p}$, $\textbf{K}^{n}_{p+1}$ and $\textbf{R}^{n}_{p+1}$ represent input knowledge concepts, input regional-visual features, output knowledge concepts and output regional-visual features at the $p^{th}$ cascade layer, respectively. For $\mathcal{F}_{cca} ^ {(1)}$, we set original input features $\textbf{R}^{n}_{1}=\textbf{R}^{n}$ and $\textbf{K}^{n}_{1}=\textbf{K}^{n}$. Finally, the outputs $[\textbf{K}^n_{P}, \textbf{R}^n_{P}]=\mathcal{F}_{cca} ^ {(P-1)}(\textbf{K}^n_{P-1}, \textbf{R}^n_{P-1})$ with Eq.~\eqref{eq:6} are regarded as the enhanced knowledge concepts and attentive regional-visual features of CCA, respectively.

The designed KAN has proved its superior potential to collect external commonsense facts and capture long-range pairwise correlations of cross-modal features, so as to better discriminate the valuable heterogeneous representations from imaginative corpus and visual contexts. Nevertheless, KAN only establishes multiple interactions of single image, neglecting to explore the topic-aware global consistency that is necessary for visual storytelling. To tackle this problem, the group-wise semantic module~(GSM) is further developed to exploit the global guided aggregation as presented in the following Section~\ref{sect:prop:gsm}.

\subsection{Group-wise Semantic Module}
\label{sect:prop:gsm}

\begin{figure}[tb]
\begin{center}
\includegraphics[width=1.0\linewidth]{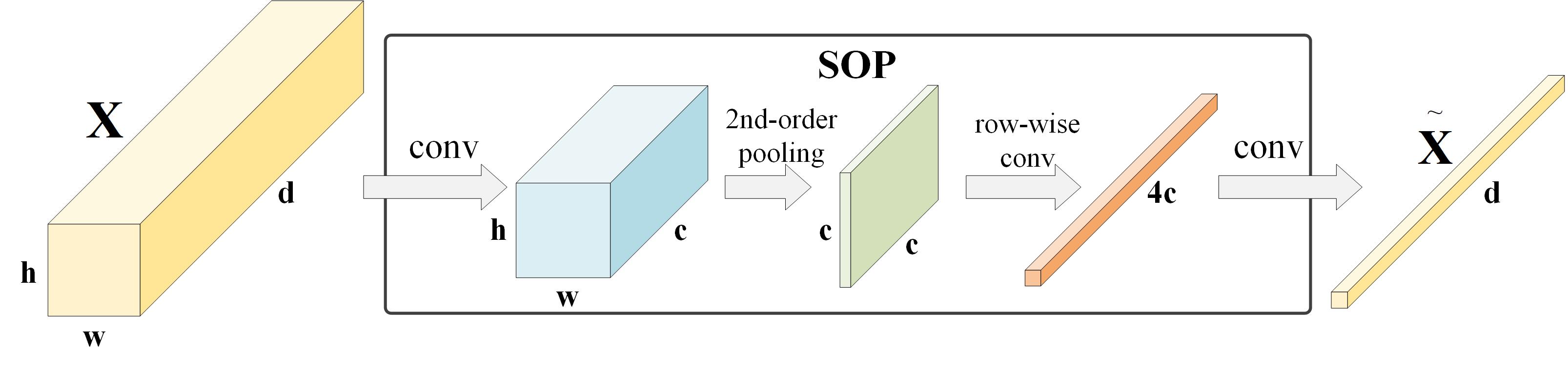}
\end{center}
\caption{The schematic diagram of SOP. Given an input feature tensor with size $h \times w \times d$, it is fed into SOP, which consists of two $1\times 1$ convolutions, one transpose multiplication operator and one row-wise convolution, generating a $1 \times 1 \times d$ global guided aggregation.}
\label{fig:attention}
\end{figure}

One major difficulty in visual storytelling task is the lack of storyline, leading to the incoherent expressions of multiple sentences. To this end, a group-wise semantic module composed of several second order pooling algorithms is developed to capture the global consistent guidance.

\noindent\textbf{$\textbf{Second Order Pooling~(SOP)}$.} Given the convolutional feature tensor $\textbf{X} \in \mathbb{R}^{h\times w \times d}$ as shown in Fig.~\ref{fig:attention},
where $h$, $w$ and $d$ represent the height, width and channel dimension of feature tensor, respectively.
SOP first introduces a $1 \times 1$ convolution to reduce the channel number from $d$ to $c$, thus projecting the convolutional feature from high to low dimension while alleviating the computation cost. Then SOP converts a $h\times w \times c$ feature tensor to a $c \times c$ covariance matrix by computing dense semantic interactions regardless of positional distance.
Each element in the covariance matrix indicates the similarity of any pairwise vectors in the feature tensor,
which formulates the high-order property of significant holistic representation by introducing a quadratic operator and thus can enable the model with the capacity of non-linear feature discrimination. Finally, a row-wise convolutional layer and a $1 \times 1$ convolutional layer are leveraged to convert the $c \times c$ covariance matrix to a $1 \times 1 \times d$ tensor to highlight the meaningful feature channels. Specifically, the process of SOP can be described as
\begin{equation}
\begin{split}
\label{eq:7}
\tilde{\textbf{X}}
&=SOP(\textbf{X})\\
&=f^{1 \times 1}(f^{row}([\mathcal{R}(f^{1 \times 1}(\textbf{X}))]^{\top}*[\mathcal{R}(f^{1 \times 1}(\textbf{X}))] )),
\end{split}
\end{equation}
where $\tilde{\textbf{X}}\in \mathbb{R} ^{1\times 1 \times d}$, $*$ indicates matrix multiplication, $\mathcal{R}$ is a reshaping operator that flattens a tensor from size $h \times w \times c$ to $(hw)\times c$, $f^{1 \times 1}$ and $f^{row}$ represent a $1\times 1$ convolution and a row-wise convolution, respectively.

\noindent\textbf{$\textbf{Group-wise Semantic}$.}  In Fig.~\ref{fig:Pipeline-acm}(c), the GSM module first inputs every feature representation $\textbf{C}^{n} \in \mathbb{R}^{h \times w \times d}$ into SOP with Eq.~\eqref{eq:7}, and then the SOP outputs the processed tensor $\tilde{\textbf{C}}^{n} \in \mathbb{R}^{1 \times 1 \times d}$. Afterwards, all processed tensors are sequentially concatenated into $\textbf{A}=[\{ \tilde{\textbf{C}}^{n} \}_{n=1} ^{N}] \in \mathbb{R} ^ {N \times 1 \times d}$, producing an initial group-wise semantic representation. Similarly, the GSM again sends $\textbf{A}$ into SOP with Eq.~\eqref{eq:7} to capture the long-range semantic associations along the channel-wise dimension, generating the global-visual aggregation $\tilde{\textbf{A}} \in \mathbb{R} ^ {1 \times 1 \times d}$ that can contribute to the subsequent multi-modal story decoder in~\secref{sect:prop:decoder}, which can be formulated as
\begin{equation}
\begin{split}
\label{eq:8}
\tilde{\textbf{A}}
&= \mathcal{F}_{gsm}(\mathcal{C})\\
&= SOP([\{ SOP (\textbf{C}^{n}) \}_{n=1} ^{N}]).
\end{split}
\end{equation}
As a consequence, the SOP can strengthen the non-linear feature capability by learning higher-order statistic dependencies of holistic representation~\cite{gao2019global}, and the GSM can capture the global consistent representation of group-wise features along the channel-wise dimension as shown in Fig.~\ref{fig:gsm}, facilitating to acquire topic-aware information for coherent and narrative descriptions.

\subsection{Multi-modal Story Decoder}
\label{sect:prop:decoder}

\begin{figure}[tb]
\begin{center}
\includegraphics[width=0.9\linewidth]{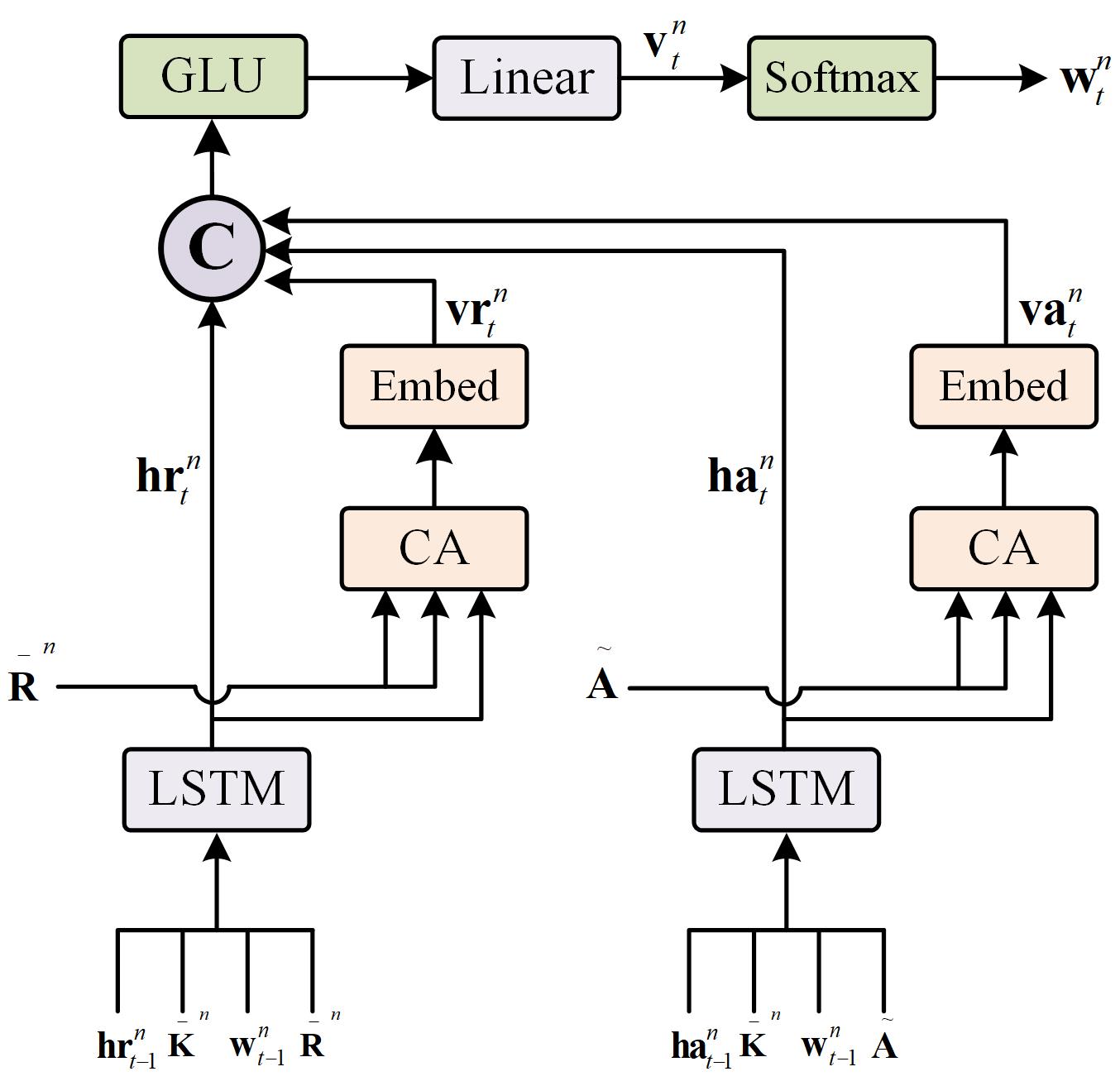}
\end{center}
\caption{Illustration of the proposed multi-modal story decoder. For the knowledge indicator vector $\bar{\textbf{K}}^{n}$, the regional-visual indicator vector $\bar{\textbf{R}}^{n}$, the global-visual indicator vector $\tilde{\textbf{A}}$, the previous regional hidden state $\textbf{hr}_{t-1}^{n}$, the previous global hidden state $\textbf{ha}_{t-1}^{n}$ and the previous word embedding $\textbf{w}_{t-1} ^{n}$ as inputs, the decoder feds these vectors into a two-stream structure by combining the CA unit and LSTM to obtain a set of vectors $\textbf{vr}_{t}^{n}$, $\textbf{hr}_{t}^{n}$, $\textbf{va}_{t}^{n}$ and $\textbf{ha}_{t}^{n}$. Finally, these vectors are concatenated and sent into following layers to obtain the current word prediction $\textbf{w}_{t}^{n}$.}
\label{fig:story-decoder}
\end{figure}

To fully utilize the produced attentive local-visual features, enhanced knowledge concepts and global-visual aggregation, a multi-modal story decoder is designed to explore the final contextual representation with above multi-modal features, generating reasonable and coherent sentences of the final story. Figure~\ref{fig:story-decoder} illustrates the diagram of the proposed multi-modal story decoder. Specifically, in order to generate the $n^{th}$ sentence with various representations including attentive regional-visual features $\textbf{R}^{n}_{P}$, enhanced knowledge concepts $\textbf{K}^{n}_{P}$ and global-visual aggregation $\tilde{\textbf{A}}$, the model first flattens $\textbf{R}^{n}_{P} \in \mathbb{R} ^{M \times d}$ to $\bar{\textbf{R}}^{n} \in \mathbb{R} ^{1 \times d}$, $\textbf{K}^{n}_{P} \in \mathbb{R} ^{K \times d}$ to $\bar{\textbf{K}}^{n} \in \mathbb{R} ^{1 \times d}$ with designed flatten layer composed of two linear layers and one softmax layer, resulting in the regional-visual indicator vector $\bar{\textbf{R}}^{n}$ and the knowledge indicator vector $\bar{\textbf{K}}^{n}$, where $M$, $K$ and $d$ denote the number of detected regional boxes, graph relations and feature channels, respectively.

To further exploit compact interactions of visual features, enriched contexts and word embedding, a regional-visual and global-visual based story decoder is designed by combining the CA unit and LSTM to accomplish multi-modal inference. Particularly, for regional-visual information reasoning of the $n^{th} $ image at the time step $t$ (see the left side of Fig.~\ref{fig:story-decoder}), the decoder sends the previous regional hidden state $\textbf{hr}_{t-1}^{n}$, the knowledge indicator vector $\bar{\textbf{K}}^{n}$, the previous word embedding $\textbf{w}_{t-1} ^{n}$ and the regional-visual indicator vector $\bar{\textbf{R}}^{n}$ into LSTM, outputting the current regional hidden state $\textbf{hr}_{t}^{n} $. Afterwards, the decoder considers $\textbf{hr}_{t}^{n} $  as the query of the CA unit, and $\bar{\textbf{R}}^{n} $ is set as the key or value of the CA unit. As a result, the output of the CA unit followed with an embedded layer obtains the attended regional representation $\textbf{vr}_{t}^{n} $  by encouraging the cross-modal correlations between  $\bar{\textbf{R}}^{n} $ and $\textbf{hr}_{t}^{n} $, which can be formulated as
\begin{align}
&\textbf{hr}_{t}^{n}= LSTM(\bar{\textbf{K}}^{n} \oplus \textbf{w}_{t-1} ^{n} \oplus \bar{\textbf{R}}^{n} , \textbf{hr}_{t-1}^{n}),\\
&\textbf{vr}_{t}^{n}= Embed (CA(\textbf{hr}_{t}^{n},  \bar{\textbf{R}}^{n} ) ),
\end{align}
where $Embed(\cdot)$ represents a fully-connected layer and $\oplus$ denotes the concatenation operator. Similarly, with the input of the previous global hidden state $\textbf{ha}_{t-1}^{n}$, the knowledge indicator vector $\bar{\textbf{K}}^{n}$, the previous word embedding $\textbf{w}_{t-1} ^{n}$ and the global-visual aggregation $\tilde{\textbf{A}}$, the global-visual information reasoning (see the right side of Fig.~\ref{fig:story-decoder}) can also generate the current global hidden state $\textbf{ha}_{t}^{n}$ and attended global representation $\textbf{va}_{t}^{n}$, which can be formulated as
\begin{align}
&\textbf{ha}_{t}^{n}= LSTM(\bar{\textbf{K}}^{n} \oplus \textbf{w}_{t-1} ^{n} \oplus  \tilde{\textbf{A}}, \textbf{ha}_{t-1}^{n}), \\
&\textbf{va}_{t}^{n}= Embed (CA (\textbf{ha}_{t}^{n},  \tilde{\textbf{A}} ) ).
\end{align}

Next, the contextual vector $\textbf{v}_{t}^{n}$ is calculated by concatenating $\textbf{vr}_{t}^{n}$, $\textbf{hr}_{t}^{n} $, $\textbf{va}_{t}^{n} $ and $\textbf{ha}_{t}^{n} $, followed with a GLU~\cite{dauphin2017language} and a linear layer, respectively. Finally, the contextual vector $\textbf{v}_{t}^{n}$ is fed into a softmax layer to generate the current word embedding $\textbf{w}_{t}^{n}$. Definitely, the word generation probability can be formulated as
\begin{equation}
\label{eq:13}
p(\textbf{w}_{t}^{n} | \textbf{w}_{1:t-1} ^{n})=softmax(\textbf{v}_{t}^{n}),
\end{equation}
where the prediction $p$ is a probability distribution over the Visual Storytelling~(VIST) dataset~\cite{huang2016visual} vocabulary $\mathbb{V}_{s}$. Finally, the word embedding $\textbf{w}_{t}^{n}$ is transformed into word $w_{t}^{n}$, obtaining the sub-story $\textbf{S}^{n}=\{ w_{1}^{n}, \cdots, w_{T}^{n} \}$ of story $\mathcal{S}$, where $T$ represents the length of sub-story $\textbf{S}^{n}$.

\subsection{Training and Inference Procedure}
\label{sect:prop:train}

In the training stage, given a group of $N$ images, all the key components of the proposed model in Fig.~\ref{fig:Pipeline-acm} are jointly trained on the VIST dataset~\cite{huang2016visual} for story prediction. The cross-entropy loss is employed in the training stage as
\begin{equation}
\label{eq:14}
L(\theta)=- \sum _{n} ^{N} \sum _{t} ^{T} log (p_{\theta}^{n} (\textbf{g}_{t} ^{n} | \textbf{g}_{1} ^{n} ,\cdots, \textbf{g}_{t-1} ^ {n})),
\end{equation}
where $\theta$ indicates the set of optimized parameters during training, $\textbf{g}_{t}^{n}$ represents the $t^{th}$ word embedding in the ground-truth sub-story $\textbf{g}^{n}$. Eventually, the goal is to minimize the loss $L(\theta)$. In the inference stage, the model predicts the story using the beam search method with the beam size equal to 3.

\section{Experiments}
\label{sect:exp}

\subsection{Implementation Details}
\label{sect:exp:impl}

Following the previous works~\cite{wang2018no,huang2019hierarchically, xu2021imagine}, the proposed KAGS model adopts the ResNet-152~\cite{he2016deep} pretrained on the ImageNet~\cite{deng2009imagenet} dataset for convolutional feature extraction and utilizes the Faster-RCNN~\cite{ren2016faster} pretrained on the ImageNet~\cite{deng2009imagenet} dataset and the Visual Genome~\cite{krishna2017visual} dataset for regional-level feature extraction, where the original convolutional feature and the regional feature are a $7\times 7 \times 2048$ tensor and a $1 \times 2048$ tensor, respectively. Then these features are transposed into tensors with the channel dimension equal to $1024$. The number of images in an album is set as $5$. For each commonsense graph, the max number of relations is set as $20$. Moreover, the number of the detected regional boxes is set as 36, the dimension of word embedding is set as $1024$, the feature dimension in the hidden layer of LSTM is set to $512$, and the number of cascade layers (\emph{i.e.}, $P-1$) in CCA is set as $6$. In the current work, the cross-entropy loss is used to train the whole model and the Adam optimizer~\cite{kingma2014adam} is employed with the initial weight decay $5\times e ^{-4}$ and the learning rate $4 \times e ^{-4}$. The model is converged in only $21$ epochs with the batch size equal to $50$, note that the model does not leverage any post processing such as reinforcement learning~\cite{rennie2017self}. The words appearing more than $3$ times in the training dataset are selected to build a storytelling vocabulary with a size of $9,837$. Then the vocabulary size is extended to $12,322$ with external knowledge base. During inference, the beam search strategy is leveraged with the beam size of $3$ for visual storytelling prediction. The model is implemented with PyTorch\footnote{[Online]. Available: https://pytorch.org/} with a Tesla V100 for acceleration.

\begin{figure*}[htpb]
\begin{center}
\includegraphics[width=0.95\linewidth]{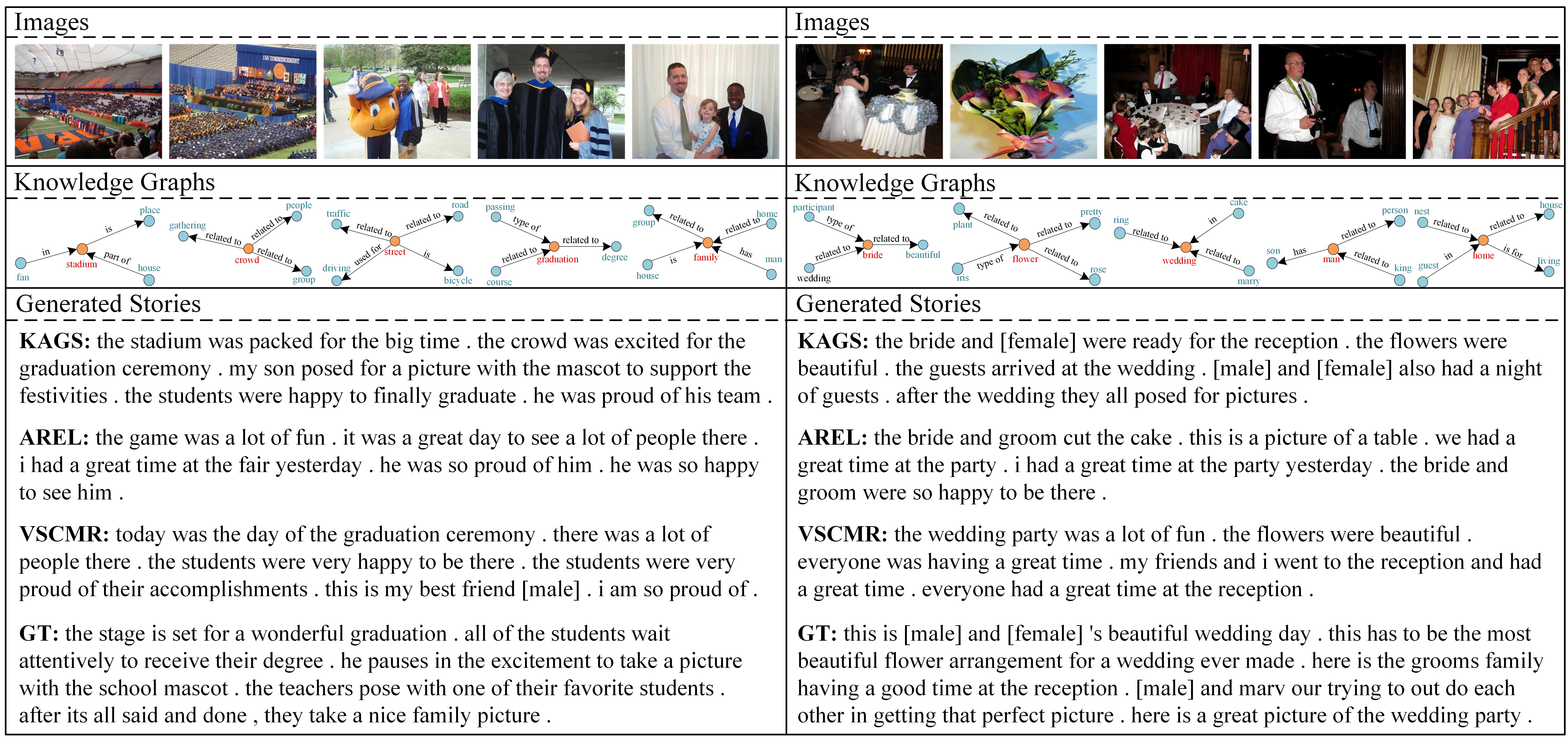}
\end{center}
   \caption{Visualization of the comparison between the proposed KAGS and other state-of-the-art methods including AREL, VSCMR and ground-truth. It only visualizes parts of the extracted commonsense knowledge graphs due to space limit.}
\label{fig:visual-compare}
\end{figure*}

\subsection{Dataset and Automatic Metric Evaluation}
\label{sect:exp:data}

\noindent\textbf{$\textbf{VIST Dataset}$.} The VIST dataset~\cite{huang2016visual} is a customized dataset for visual storytelling, which contains $210,819$ specific images and $10,117$ interesting Flicker albums. It is challenging to employ VIST for visual storytelling, because the story descriptions are more subjective and need emotional and imaginative concepts that do not appear explicitly in images. Following the previous work~\cite{wang2018no}, the broken photos are removed and $40,098$ training groups, $4,988$ validation groups and $5,050$ testing groups are constructed. Each group consists of $5$ images collected from one photo album and each image usually corresponds to one sentence. Every album has $5$ differentiate stories as reference.

\noindent\textbf{$\textbf{Automatic Metric Evaluation}$.} Comprehensive experiments are conducted on the VIST dataset in terms of four automatic metrics including BLEU~\cite{papineni2002bleu}, METEOR~\cite{banerjee2005meteor}, ROUGE\_L~\cite{lin2004rouge} and CIDEr~\cite{vedantam2015cider}. These metrics calculate the similarities and relevances between the predicted story and reference. Concluding in~\cite{huang2016visual}, the METEOR score is chosen as the key performance indicator for its high correlation with human evaluation standards.

\subsection{Comparison with State-of-the-art Methods on Automatic Metrics}
\label{sect:exp:sota}

The proposed KAGS model is compared with other twelve state-of-the-art visual storytelling approaches including (1) seq2seq~\cite{huang2016visual}, an original model with RNN-based structure; (2) BARNN~\cite{liu2017let}, a relational attended model with designed GRU; (3) h-attn-rank~\cite{yu2017hierarchically}, a hierarchical attentive recurrent network; (4) XE-ss~\cite{wang2018no}, a LSTM-based encoder-decoder model; (5) AREL~\cite{wang2018no}, an adversarial reward optimizing framework; (6) HPSR~\cite{wang2019hierarchical}, a hierarchical image encoder-decoder model; (7) HSRL~\cite{huang2019hierarchically}, a hierarchical reinforcement learning framework; (8) VSCMR~\cite{li2019informative}, a conceptual exploration network; (9) ReCO-RL~\cite{hu2020makes}, a relevant context reinforcement learning method; (10) INet~\cite{jung2020hide}, an imaginative concept reasoning network; (11) SGVST~\cite{wang2020storytelling}, a scene-graph knowledge enhanced model; and (12) IRW~\cite{xu2021imagine}, a multi-graph knowledge reasoning framework. For fair comparisons, this paper directly presents the statistic results provided by the authors or conducts the experiments by the official source codes of these competing approaches.

\subsubsection{Qualitative Results}
\label{sect:exp:sota:qlres}

Figure~\ref{fig:visual-compare} presents several visual comparisons between the proposed KAGS model and the methods AREL~\cite{wang2018no} and VSCMR~\cite{li2019informative}, together with the human-annotated referenced stories~(GT). Generally, compared with the other two approaches (\emph{i.e.}, AREL and VSCMR), KAGS can better generate emotional, imaginative, coherent and accurate descriptions by jointly exploring the knowledge enriched cross-modal interactions and global semantic guidance.

Specifically, the left album of the five images in Fig.~\ref{fig:visual-compare} is related to a graduation activity with various scenes, it is apparent that the predicted sentences obtained by KAGS show promising performances. For the second sentence of VSCMR, it simply produces the sentence ``there was a lot of people there'' and neglects to record the detailed visual and implicit contexts in this picture, leading to sub-optimal results. Notwithstanding, the second sentence of KAGS shows the description ``the crowd was excited for the graduation ceremony'', where the word ``excited'' properly depicts the emotions of people and the phrase ``graduation ceremony'' accurately illustrates the social activity using the information from knowledge graphs, confirming the capability of KAGS to capture rich emotions and external contexts according to visual and textual information. For the fourth sentence of AREL (\emph{i.e.}, ``he was so proud of him''), it only characterizes the emotions of people and is irrelevant to the precise visual context in this photo, which is ambiguous to understand. However, the fourth sentence of KAGS outputs the sentence ``the students were happy to finally graduate'', which highly corresponds to the graduation topic of this photo album.

\begin{table*}[htbp]
  \caption{ Statistic comparisons of KAGS with other state-of-the-art approaches, where the bold font indicates the best performance.
  }\vspace{-8pt}
  \centering
  \renewcommand{\arraystretch}{1.2}
  \renewcommand{\tabcolsep}{9pt}
  \normalsize
  \begin{tabular}{l||ccccccc} \hline \toprule
    Methods& BLEU-1  & BLEU-2 & BLEU-3 & BLEU-4  & METEOR &ROUGE\_L & CIDEr  \\ \hline
seq2seq~\cite{huang2016visual} (NAACL2016)          &  - &  - & -  & 3.5  & 31.4 &  - &  6.8\\
BARNN~\cite{liu2017let} (AAAI2017)          & -  &  - & -  & -  &  33.3 &  -&  -\\
h-attn-rank~\cite{yu2017hierarchically} (EMNLP2017)          & -  &  - & 21.0  & -  &  34.1 & 29.5&  7.5\\
XE-ss~\cite{wang2018no} (ACL2018)          &  62.3 &  38.2 &  22.5 & 13.7  &  34.8 & 29.7 &  8.7\\
AREL~\cite{wang2018no} (ACL2018)           & 63.7  &  39.0 &  23.1 & 14.0  &  35.0 & 29.6 & 9.5\\
HPSR~\cite{wang2019hierarchical} (AAAI2019)          & 61.9  & 37.8  &  21.5 &  12.2 & 34.4  & 31.2 & 8.0\\
HSRL~\cite{huang2019hierarchically} (AAAI2019)          & -  & - &  - &  12.3 & 35.2  & 30.8 & 10.7\\
VSCMR~\cite{li2019informative} (ACMMM2019)           &   63.8&  39.5 & 23.5  & 14.3  & 35.5  & 30.2 & 9.0\\
ReCO-RL~\cite{hu2020makes} (AAAI2020)           &  - &  - & -  & 12.4  & 33.9  & 29.9 & 8.6\\
INet~\cite{jung2020hide} (AAAI2020)           &  64.4 &   40.1&   23.9&   14.7&  35.6 &  29.0 & 10.0\\
SGVST~\cite{wang2020storytelling} (AAAI2020)     &  65.1 &   40.1&  23.8 &   14.7 &   35.8  & 29.9 & 9.8\\
 IRW~\cite{xu2021imagine} (AAAI2021)        & 66.7  &   41.6 & 25.0  &  \textbf{15.4}&   35.6  & 29.6 & 11.0 \\
\hline
\textbf{KAGS}       &  \textbf{70.1} & \textbf{43.5}   & \textbf{25.2}  & 14.7  & \textbf{36.2}  & \textbf{31.4} & \textbf{11.3 }\\
\hline \toprule
\end{tabular}
\label{tab:comparision}
\end{table*}

Moreover, the story generation of the right photo album in Fig.~\ref{fig:visual-compare} is also challenging due to its numerous characters and various semantic objects in different scenarios. In the estimated story obtained by AREL, the third and fourth sentences show the repetitive phrase ``had a great time'', which impairs the abundant descriptions of this story. Notwithstanding, the proposed KAGS can avoid this problem and generate sentences with different formats and styles (\emph{i.e.}, the third and fourth sentences generated by KAGS). In addition, regarding the fifth sentence of story, the VSCMR method predicts the sentence of ``everyone had a great time at the reception'', which generally introduces the event happened in this scene. And the proposed KAGS generates the sentence of ``after the wedding they all posed for pictures'', which shows that the generated sentence is associated with the visual information in the fourth image, further validating the long-range dependency capacity of the proposed KAGS model. Totally, the experimental results demonstrate that the designed model is able to obtain favorable story estimations in several challenging conditions, confirming the superior performance of the proposed KAGS model.

\subsubsection{Quantitative Results}
\label{sect:exp:sota:qnres}

The comparison of the proposed KAGS model with other state-of-the-art approaches is also presented in Table~\ref{tab:comparision}, where it can be observed that the statistic results of KAGS show better performances than the competing approaches by a large margin. Generally, the proposed KAGS achieves the best scores in terms of six metrics including BLEU-1, BLEU-2, BLEU-3, METEOR, ROUGE\_L and CIDEr, and obtains the second best performance on BLEU-4. Specifically, KAGS achieves the BLEU-1 score of 70.1, the BLEU-2 score of 43.5, the BLEU-3 score of 25.2, the METEOR score of 36.2, the ROUGE\_L score of 31.4 and the CIDEr score of 11.3, significantly surpassing the scene graph based method SGVST~\cite{wang2020storytelling} by $5.0\%$, $3.4\%$, $1.4\%$, $0.4\%$, $1.5\%$ and $1.5\%$, respectively. Moreover, compared with the second best method IRW~\cite{xu2021imagine} that leverages many external knowledge including scene graph, commonsense graph and event graph, the proposed KAGS model can achieve higher scores on most metrics. Particularly, 70.1 versus 66.7 on BLEU-1, 43.5 versus 41.6 on BLEU-2, 25.2 versus 25.0 on BLEU-3, 36.2 versus 35.6 on METEOR, 31.4 versus 29.6 on ROUGE\_L, 11.3 versus 11.0 on CIDEr. In summary, the quantitative results confirm that the proposed modules can boost the performance of visual storytelling by enhancing interactions of heterogeneous information and capturing the global guidance of storyline.

\subsection{Experimental Analysis}
\label{sect:exp:aly}

\subsubsection{Ablation Study}
\label{sect:exp:aly:abl}

To investigate the effectiveness of the proposed modules, ablative experiments are conducted in absence of KAN $\&$ GSM (KAGS-KG), KAN (KAGS-K), CCA (KAGS-C) and GSM (KAGS-G), respectively. The statistic results are presented in Table~\ref{tab:ablation}.

\begin{table}[htbp]
\vspace{-8pt}
\renewcommand{\arraystretch}{1.2}
\renewcommand{\tabcolsep}{1.5pt}
\caption{Ablation study of the proposed model on the VIST dataset, here KAGS-KG, KAGS-K, KAGS-C and KAGS-G represent the model without KAN $\&$ GSM, KAN, CCA, and GSM, respectively. The bold font represents the best performance.}
\normalsize
\begin{tabular}{l||c|c|c|c|c} \hline \toprule
Metrics& KAGS-KG & KAGS-K  & KAGS-C & KAGS-G & \textbf{KAGS}   \\ \hline
BLEU-1           &  62.6 &  66.7 & 68.5 &68.4 & \textbf{70.1}    \\
BLEU-2          &  37.7 &   41.7  & 42.5 &42.0  & \textbf{43.5}  \\
BLEU-3          &  21.6 &  24.4  & 24.8& 24.7 & \textbf{25.2}    \\
BLEU-4          & 12.7 &   14.4  & 14.6 &14.5 & \textbf{14.7 }   \\
METEOR         & 34.3 &  36.0  & 35.5&35.4  & \textbf{36.2 }     \\
ROUGE\_L           & 28.5 &  31.2  & 30.5&31.1 & \textbf{31.4 }     \\
CIDEr             &  7.8  &  9.5  & 11.0& 10.7  & \textbf{11.3}  \\
\hline \toprule
\end{tabular}
\label{tab:ablation}
\end{table}

\begin{figure}[htbp]
\begin{center}
\includegraphics[width=0.95\linewidth]{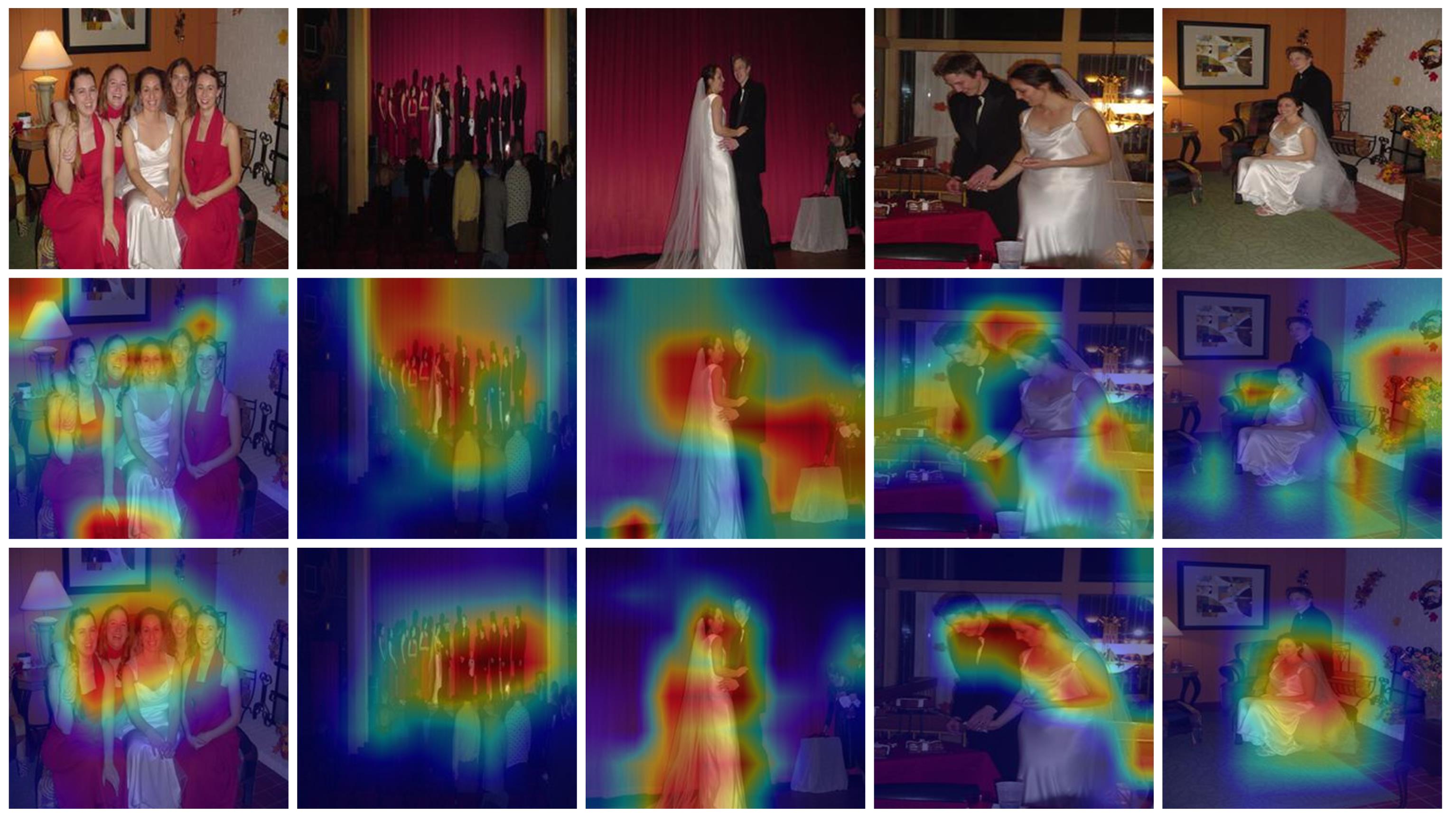}
\end{center}
\caption{Effectiveness of GSM to capture topic-aware global consistency. Top to down: input images, activation maps without GSM, activation maps with GSM.}
\label{fig:gsm}
\end{figure}

First, without GSM, the KAGS-G presents apparent performance degradation on the VIST dataset, particularly on the metrics of BLEU-1 and BLEU-2 with the evaluation scores being declined from 70.1 to 68.4 by $1.7\%$, from 43.5 to 42.0 by $1.5\%$, respectively. In addition, the visualized activation maps obtained by GSM are illustrated in Fig.~\ref{fig:gsm}, which proves that GSM can focus more attention on the global consistent regions while removing the semantic foreground and background interferences. Therefore, the statistic results prove the positive effects of GSM to capture the long-range dependencies for global guidance.

Second, without CCA, the statistic results of KAGS-C also show obvious performance drop on all metrics, especially on the metrics of METEOR and ROUGE\_L, the former score reduces from 36.2 to 35.5 by $0.7\%$ and the latter score reduces from 31.4 to 30.5 by $0.9\%$, respectively. The ablative results verify the effectiveness of the designed CCA to establish the cross-modal interactions for visual and textual information enhancement.

Third, without KAN, all the metrics obtained by KAGS-K present significant decrease on the VIST dataset. Especially, KAGS outperforms KAGS-K by a large margin in terms of BLEU-1, BLEU-2 and CIDEr, with the scores being 70.1 versus 66.7, 43.5 versus 41.7 and 11.3 versus 9.5, respectively. It is worth noting that KAN can capture the external rich knowledge and explore the correlation of heterogeneous information, facilitating to more abundant and reasonable descriptions.

Finally, without KAN and GSM, the statistic performance of KAGS-KG has extreme decline in terms of all metrics, further demonstrating the superiority of the designed KAN and GSM to learn attentive multi-modal representation and global semantic tailored to the visual storytelling task.

\subsubsection{Visualization Analysis}
\label{sect:exp:aly:vis}

\begin{figure*}[htbp]
\begin{center}
\includegraphics[width=0.95\linewidth]{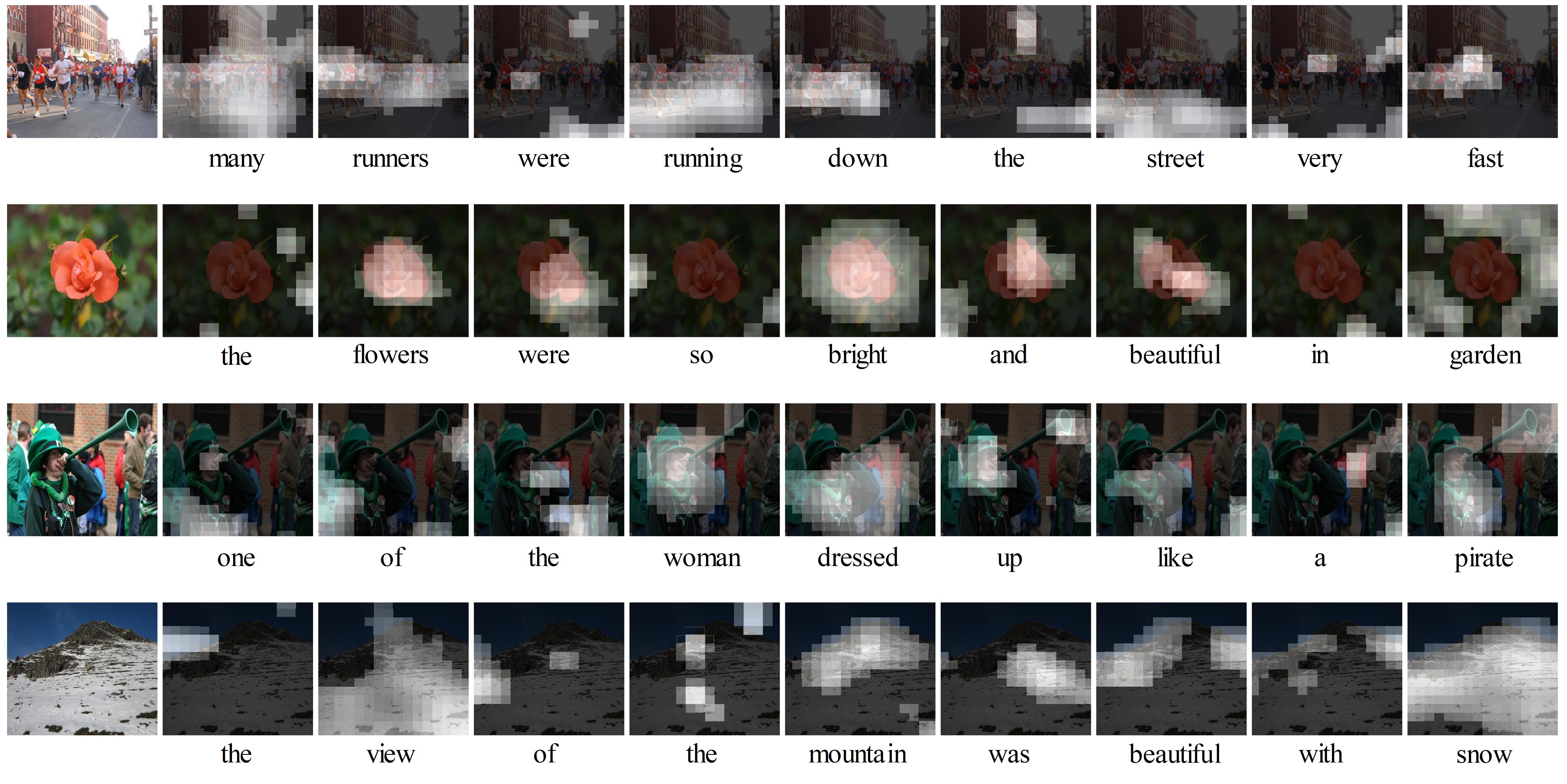}
\end{center}
\caption{Visualization of KAN, where the whiter color of an image area represents that higher attention weights are given to that area.}
\label{fig:kan}
\end{figure*}

In order to better verify the effectiveness of GSM and KAN, the class activation map~\cite{zhou2016learning} of each image and the attention distributions of each image region during word generation are visualized in Fig.~\ref{fig:gsm} and Fig.~\ref{fig:kan}, respectively.

First, as aforementioned, the class activation map of each image is visualized in Fig.~\ref{fig:gsm}, where the class activation map is computed by $\textbf{M} ^{n} = \textbf{C}^{n} \tilde{\textbf{A}} ^\top$ referenced from ~\cite{zhou2016learning}. In the second line of Fig.~\ref{fig:gsm}, the model fails to discriminate the consistency among group images and suffers from the background clutters, such as wrongly localizing the people under the stage in the second image and introducing the background interferences in the third image. Nevertheless, the designed GSM can well capture the consistent characteristics of bride and groom in this image sequence and suppress the background clutters, thus again confirming the advantages of triggering the global semantic of group-wise features.

Second, several generated sentences of differentiate images are presented in Fig.~\ref{fig:kan} to illustrate the effectiveness of KAN, where the whiter the color of image regions are, the higher attention weights are given to these regions. When referring to generate the nouns (\textit{e.g.}, `runners', `street', `flowers', `woman', `mountain'), the module prefers to assign higher weights to the relevant areas; when predicting the verbs, KAN often gives more valuable attention weights to both of the local and non-local areas of relative action. Moreover, the imaginative words can be assigned with higher attention scores by KAN according to surrounding environments. For example, in the first line of Fig.~\ref{fig:kan}, the region corresponding to the noun `runners' is highlighted by assigning higher attention weights, when generating the verb `running', the module pays more attention on runners' legs as well as their whole bodies. In the second line of Fig.~\ref{fig:kan}, the noun `garden' doesn't significantly appear in this image, but higher weights are correctly assigned to the surrounding areas of the flower. The visualized examples further verify the merit of KAN of paying attention to important regions, meaningful actions and abstract areas.
\begin{table*}[htbp]
\caption{Statistic results of human evaluation metrics, here the percentage numbers represent the confident scores of the tester believe that a model surpasses its opponent, and Tie means the tester can not choose the better story.}
\vspace{-8pt}
\centering
\renewcommand{\arraystretch}{1.2}
\renewcommand{\tabcolsep}{2.4pt}
\normalsize
\begin{tabular}{l||c |c |c | c |c |c | c |c |c | c |c |c} \hline \toprule
Methods     & \multicolumn{3}{c|}{XE-ss vs KAGS } & \multicolumn{3}{c|}{AREL vs KAGS } & \multicolumn{3}{c|}{VSCMR vs KAGS} & \multicolumn{3}{c}{IRW vs KAGS} \\ \hline
Choice             & XE-ss & KAGS & Tie & AREL & KAGS& Tie & VSCMR & KAGS & Tie& IRW &  KAGS & Tie \\ \hline
Relevance & 35.9\% & \textbf{59.5\%} & 4.6\% & 38.2\% & \textbf{51.0\%} & 10.8\% & 32.1\% & \textbf{47.6\%} & 20.3\% & 36.0\% & \textbf{42.9\%} & 21.1\%\\
Expressiveness & 27.1\% & \textbf{66.4\%} &6.5\% & 31.5\%& \textbf{58.8\%} & 9.7\% & 33.5\%    & \textbf{45.2\%} & 21.3\% & 34.3\% & \textbf{39.6\%} & 26.1\%\\
Concreteness & 32.8\% & \textbf{60.9\%} & 6.3\%& 37.9\%& \textbf{49.4\%} & 12.7\%& 30.8\%    & \textbf{44.3\%} & 24.9\% & 31.7\% & \textbf{37.2\%} & 31.1\%\\
\hline \toprule
\end{tabular}
\label{tab:human-comparison}
\end{table*}
\subsection{Human Evaluation}
\label{sect:exp:hueva}

The previous works~\cite{wang2018no,li2019informative} have concluded that automatic evaluation metrics can not reflect the semantic properties of many stories (\textit{e.g.}, coherence and expressiveness), therefore human evaluation metrics~\cite{li2019informative} are further adopted for comparison in pairwise manner. Specifically, $150$ photo albums with a total of $750$ images from the VIST test dataset are randomly selected and two stories generated by KAGS and another competing method are presented for every volunteer, noting that the optional orders in each item are shuffled for fairness. Then each volunteer needs to choose a better story according to the metrics of relevance, expressiveness and concreteness. The detailed illustrations of these three criteria are defined as follows.
\begin{itemize}
\item
\textbf{Relevance} describing the precise topic of happened activity in image sequence.
\item
\textbf{Expressiveness} generating the grammatical, imaginary, coherent and abundant sentences.
\item
\textbf{Concreteness} providing the narrative and concrete descriptions of image contexts.
\end{itemize}

Table~\ref{tab:human-comparison} lists four comparison tests: XE-ss~\cite{wang2018no} \textit{vs} KAGS, AREL~\cite{wang2018no} \textit{vs} KAGS, VSCMR~\cite{li2019informative} \textit{vs} KAGS, and IRW~\cite{xu2021imagine} \textit{vs} KAGS. As seen from the results, it is obvious that the statistic results of KAGS are better than other competing methods in all the three metrics. Especially, the scores of KAGS are much higher than XE-ss by $23.6\%$, $39.3\%$ and $28.1\%$ in terms of relevance, expressiveness and concreteness, respectively. Compared with newest method IRW, the proposed KAGS model also shows superior performances and achieves more significant advantages than the scores on automatic evaluation metrics. Thus, it can empirically prove that the generated stories of KAGS can better obtain the storyline of image sequence, produce the imaginative words and generate concrete descriptions, which can not be obviously revealed by automatic metrics.

\section{Conclusion}
\label{sect:cln}

A knowledge-enriched attention network with group-wise semantic for visual storytelling has been developed, which consists of two main novel designs: KAN and GSM. The proposed KAN is designed to leverage the external knowledge and visual information extracted to characterize the cross-modal interactions with attention mechanism. In order to obtain the storyline with global feature guidance, a novel GSM is devised to explore the group-wise semantic with second-order pooling. All these extracted multi-modal representations are then fed into the decoder for story generation. Finally, a one-stage encoder-decoder framework is established to optimize all these designed modules in an end-to-end manner. Extensive experiments on the VIST dataset have been carried out to demonstrate the superior performance of the proposed KAGS model as compared with other state-of-the-art methods. The proposed KAGS scheme is capable of learning robust feature representations at regional and global levels to achieve superior performances. However, there is still some gaps between the storyline generated by KAGS and that of human storytellers who are trained to generate narrative stories with human language styles. We are working on taking this KAGS to its next level by considering the following three aspects: (1) investigating reinforcement learning rewards correlated with human evaluation to enhance natural expression, (2) studying more effective frameworks to accomplish visual storytelling in more sophisticated and realistic scenarios which contain much interference, and (3) generating dense visual storytelling under a complex scenario where the target image sequence contains multiple storylines.

{\small
\bibliographystyle{IEEEtran}
\bibliography{ref}
}

\vfill


\end{document}